\documentclass[letterpaper, 10 pt, conference]{ieeeconf}  

\IEEEoverridecommandlockouts                              

\usepackage{caption}
\usepackage{float}
\usepackage{multirow}
\usepackage{booktabs}
\usepackage{subcaption}
\usepackage{xcolor}
\usepackage{graphics} 
\usepackage{epsfig} 
\usepackage{mathptmx} 
\usepackage{times} 
\usepackage{amsmath,amssymb,amsfonts}
\usepackage{bm}
\usepackage{todonotes}

\makeatletter
\let\NAT@parse\undefined
\makeatother
\usepackage[colorlinks,citecolor=red,urlcolor=blue,bookmarks=false,hypertexnames=true]{hyperref}
\hypersetup{
    urlcolor=cyan,
    pdftitle={eWand},
    pdfpagemode=FullScreen,
    }
\usepackage{cleveref}

\title{\LARGE \bf eWand: An extrinsic calibration framework for wide baseline frame-based and event-based camera systems}

\author{Thomas Gossard$^{*}$, Andreas Ziegler$^{*}$, Levin Kolmar,  Jonas Tebbe and Andreas Zell
\thanks{$^{*}$ Equal contribution}
\thanks{The authors are with the Cognitive Systems Group, Dept. Informatics, University of Tuebingen. Corresponding author thomas.gossard@uni-tuebingen.de.}
\thanks{This work was funded by Sony AI.}}

\begin{document}

\maketitle
\thispagestyle{empty}
\pagestyle{empty}


\begin{abstract}
Accurate calibration is crucial for using multiple cameras to triangulate the position of objects precisely.
However, it is also a time-consuming process that needs to be repeated for every displacement of the cameras.
The standard approach is to use a printed pattern with known geometry to estimate the intrinsic and extrinsic parameters of the cameras.
The same idea can be applied to event-based cameras, though it requires extra work.
By using frame reconstruction from events, a printed pattern can be detected.
A blinking pattern can also be displayed on a screen.
Then, the pattern can be directly detected from the events.
Such calibration methods can provide accurate intrinsic calibration for both frame- and event-based cameras.
However, using 2D patterns has several limitations for multi-camera extrinsic calibration, with cameras possessing highly different points of view and a wide baseline.
The 2D pattern can only be detected from one direction and needs to be of significant size to compensate for its distance to the camera.
This makes the extrinsic calibration time-consuming and cumbersome.
To overcome these limitations, we propose \textbf{eWand}, a new method that uses blinking LEDs inside opaque spheres instead of a printed or displayed pattern.
Our method provides a faster, easier-to-use extrinsic calibration approach that maintains high accuracy for both event- and frame-based cameras.
\end{abstract}

\section*{Supplementary Material}
The project’s code and additional resources are available at: \url{https://cogsys-tuebingen.github.io/ewand/}

\section{Introduction}
\label{sec:intro}

Before using cameras for robotic applications, one needs to calibrate them first.
Camera calibration is the process of estimating the intrinsic camera parameters (the camera matrix $\bm{K}$ and the distortion coefficients $\bm{d}$) as well as the extrinsic camera parameters (the rotation $\bm{R}$ and translation $\bm{t}$ of the camera with respect to a reference frame) when multiple cameras are used.
Thus, after calibrating a camera system, we know how a point in the 3D space will be projected into the 2D captured images.
This enables the use of computer vision algorithms like triangulation or Perspective-n-Point. 

Camera calibration for conventional cameras is a well-studied subject, with literature published over multiple decades~\cite{Clarke1998tpr}\cite{ZhengyouZhang1999iccv}\cite{Rehder2016icra}.
The current best practice is to capture images of a printed pattern in various positions and orientations.
The most popular patterns are the checkerboard, the asymmetrical circle grid, and AprilTags.
\begin{figure}
    \centering
    \includegraphics[width=0.9\linewidth]{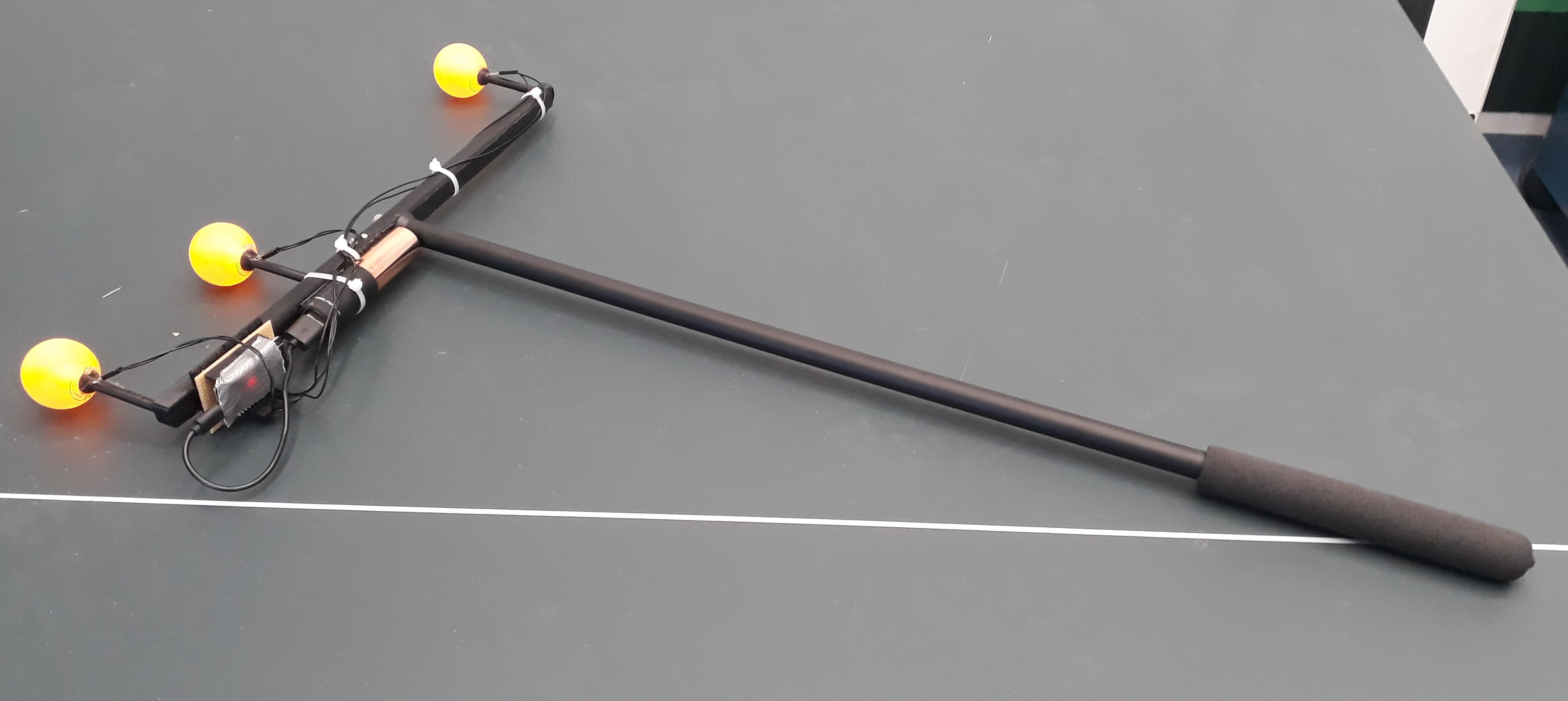}
    \caption{Our eWand, composed of three of our markers.}
    \label{fig:ewand}
\end{figure}
Calibration with such patterns work fine for a setup with a single camera or with cameras with a very similar field of view, e.g., stereo-vision setups.
However, calibrating multiple cameras with different viewpoints can be problematic for two reasons.
First, capturing images where the pattern is recognized by all cameras is not always possible, requiring calibration of cameras by pairs.
When optimizing camera pairs individually, the pairwise optimal solution often does not coincide with the optimal solution using all cameras combined.
Second, the pattern often has to be significantly tilted to be visible by multiple cameras, leading to difficulties in pattern detection and might lower the accuracy of the marker localization, potentially resulting in an inaccurate calibration.
This is the reason why for such camera setups, 3D markers (such as spheres) are preferred instead of 2D markers.
Sphere markers are always identically viewed, independent of the point of view.
Commercial motion capture tools such as Optitrack\footnote{\url{https://optitrack.com/}} or Vicon\footnote{\url{https://www.vicon.com/}} use "wands" to calibrate their cameras.
Their infrared (IR) cameras capture the positions of IR-reflective spheres mounted on a stick.
Though both Optitrack and Vicon are the industrial standard in terms of accuracy for motion capture, these systems are also quite expensive. Moreover, they require extra steps if they are to be used to calibrate a setup with standard RGB cameras. 
This concept has been translated to RGB cameras, using colored spheres and color filters to extract the position of the spheres \cite{mitchelson2003surrey}.
However, such wands cannot be directly used for event-based cameras for calibration.

Event-based cameras are a new type of camera that have known rapid growth and increase in popularity in recent years.
Instead of synchronously capturing all pixel values at periodic intervals, like frame-based cameras, event-based cameras continuously and asynchronously emit a signal (so-called \textit{events}) on a log-intensity change above a threshold for every pixel independently.
Because of their low latency, high dynamic range, and non-redundant information flow, they offer many benefits for robotic applications~\cite{Gallego2020pami}.
They have already been used for stereo-vision~\cite{Zhou2018eccv}, SLAM~\cite{HidalgoCarrio2022cvpr} and feature tracking~\cite{Chiberre2021cvprw}.

As with frame-based cameras, event-based cameras also need to be calibrated first.
The same calibration techniques as for conventional cameras cannot be directly applied.
There are two approaches for event-based camera calibration: either frames are reconstructed from the events and then traditional calibration methods can be used~\cite{Muglikar2021cvprw} or the events are directly used~\cite{Huang2021iros}\cite{Salah2023arxiv}.

Performing extrinsic calibration for frame-based and event-based cameras \textit{jointly} requires having the pattern be detectable by both types of cameras \textit{simultaneously}.
The printed pattern and the LCD screen are the common approaches.
The printed pattern is directly observable by the frame-based cameras and needs to be reconstructed for the event-based cameras.
Depending on the camera used, the LCD screen can be set to either blinking mode or not.  
However, both also rely on 2D markers, which makes them inappropriate for a setup with multiple cameras with very different points of view.
Moreover, for camera systems with a wide baseline between cameras, the pattern needs to be of significant size to be robustly detected from all cameras.
It is obvious that a large LCD screen is heavy and difficult to move around.
But this is also the case for printed patterns.
The pattern's flatness requirement necessitates a rigid and consequently heavy support to prevent bending.
This makes moving the pattern around difficult and the extrinsic calibration process slow.
Wand-based calibration approaches are therefore preferred for multi-camera setups with very different points of view and wide baselines.

In order to make the wand detected by both, the frame-base and event-based cameras, we propose the use of blinking LEDs as markers.
However, it isn't easy to pinpoint the center of an LED.
Their small size and non-spherical shape make their detection and center regression difficult.
Additionally, due to the high luminosity of the LED compared to the rest of the scene, smear, or blooming might be observed with the frame-based cameras, which makes it further challenging to estimate the center of the LED.
Therefore, LED patterns tend to have a higher reprojection error~\cite{Muglikar2021cvprw}.
To overcome these shortcomings, we propose placing these LEDs inside colored, opaque spheres.
The opaque spheres act as a diffuser and are also easier to detect due to their size and symmetry. 
This enables both event and frame-based cameras to detect the balls.
We name our method \textbf{eWand}.
Using \textbf{eWand}, we can capture points over all cameras and use them to calculate the cameras' extrinsic parameters.

\textbf{Contributions} of this work are as follows:
\begin{itemize}
    \item Robust and accurate markers for frame- and event-based cameras
    \item A detection method for said markers
    \item Extending the wand-based extrinsic calibration to event-based cameras
\end{itemize}

\section{Related Work}

Since calibration of conventional cameras with printed calibration targets is a well-established topic with several toolboxes freely and commercially available~\cite{Rehder2016icra}\cite{ZhengyouZhang1999iccv}\cite{tangram_calibration}, we will focus on the calibration of event-based cameras and wand-based calibration methods in the related work.

\subsection{Event-camera calibration}
As already mentioned, there are two methodologies for event-based camera calibration:
Either frames are reconstructed from the event stream, and then traditional calibration methods can be used, or the events are directly used for the calibration.
The first approach was introduced by~\cite{Muglikar2021cvprw}.
In this work, E2VID~\cite{Rebecq2019pami} is used to reconstruct frames from the event stream.
The reconstruction is a quite computationally heavy step (max frame rate of 11Hz for a resolution of 1280x720 with a NVIDIA RTX 2080 Ti GPU~\cite{Rebecq2019pami}).
More recent reconstruction networks such as FireNet\cite{Scheerlinck2020wacv} provide faster reconstruction, being approximately three times faster than E2VID.
However, even with reconstructed frames, pattern detection can be challenging.
Even more so with a tilted pattern, which is often the case for patterns observed from highly different points of view.
With the reconstructed frames, the kalibr calibration toolbox~\cite{Rehder2016icra} is used to calibrate in the same way as for conventional cameras.

The second approach uses the events directly without any reconstruction for faster calibration and higher detection rates of the patterns.
Early works make use of active markers such as blinking LED patterns~\cite{rpg_calibration}\cite{DominguezMorales2019ieee} or blinking screens~\cite{Mueggler2014iros}\cite{vlo_calibration}\cite{metavision_calibration} to detect and extract the calibration patterns.
In a more recent work~\cite{Huang2021iros}, the authors introduced a dynamic event-based camera calibration algorithm.
The approach directly calibrates given events captured during relative motion between the camera and the calibration pattern.
However, the approach is not designed to calibrate multiple event-based cameras extrinsically.
Although the calibration approach presented in~\cite{Huang2021iros} does use a circular pattern, often used for calibration of conventional cameras, it is not straight forward to calibrate event-based and frame-based cameras simultaneously.
Another intrinsic calibration method was later introduced in~\cite{Salah2023arxiv}.
Their approach is similar to~\cite{Huang2021iros} with improvements such as better clustering performance and achieving sub-pixel localization accuracy of the circular pattern.

Most of these methods could also be used for joint frame- and event-based camera calibration, if the event-based pattern extraction outputs the pattern in a data structure which can later be optimized together with the pattern extracted from the frame-based cameras.
Reconstruction-based methods already rely on a pattern that can be directly used by the frame-based cameras.
But as we explained, printed 2D patterns are not suited for calibrating multiple cameras with highly different points of view.
Printed patterns are moreover not as straightforward as methods based on active markers.
They can be set to always on for frame-based cameras or to blinking for event-based cameras. 
But the pattern then needs to be perfectly static between each capture, which makes the calibration much more time-consuming and needs more care.

\subsection{Wand-based calibration}
While the majority of the camera calibration literature and available calibration tool boxes use printed calibration targets for their ease of use, there also exist methods that use a wand as calibration target~\cite{mitchelson2003surrey}\cite{Shin2012ijpem}.
Using a wand as a calibration target allows rapid, flexible, and accurate calibration of multiple frame-based camera systems and does not require the cameras to all face in the same direction or to have overlapping fields of view~\cite{mitchelson2003surrey}.
They can be used to calibrate extrinsics and the intrinsics to some degree, as shown by the Optitrack or Vicon systems (it is, however, unclear how the focal length of the cameras is estimated).
While these methods were successfully applied for motion tracking with multiple infrared cameras, a wand-based calibration has never been tested to calibrate frame and event-based cameras jointly.

\section{Event camera marker}\label{sec:event_camera_marker}

As previously mentioned, calibration targets are crucial for the calibration of cameras.
In this section, we will describe our marker, which can be detected by frame- and event-based cameras.

\begin{figure}[ht]
    \centering
    \includegraphics[width=0.7\linewidth]{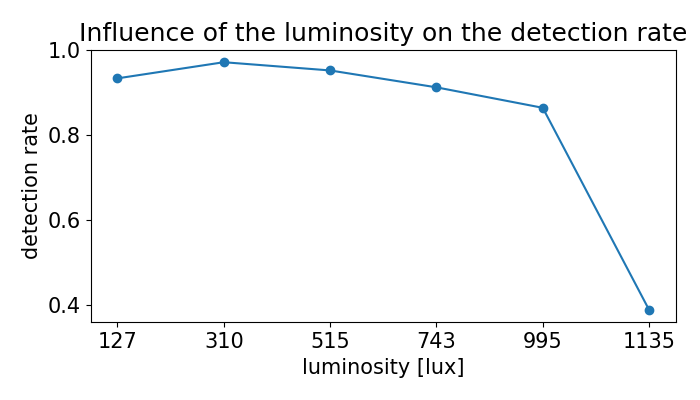}
    \caption{Influence of the luminosity on the detection rate of the marker.
            %
            }
    \label{fig:illumination_test}
\end{figure}

Blinking LEDs have already been tested and used as markers for event cameras \cite{Muglikar2021cvprw}\cite{propheseeLed}.
However, due to their shape, the identified position of the LED will depend on the point of view.
Moreover, their small size makes detecting them from long distances difficult, as they can be mistaken for noise.
We solve both of these issues by placing the LEDs in a spherical diffuser.
The spherical shape of the diffuser makes the center of the marker independent of the point of view.
This enables us to find the center of the marker more accurately.
To ensure homogeneous and powerful lighting inside the diffuser, we used 2 high-intensity LEDs (7500MCD) oriented orthogonally.
We tested the detection of our marker for different ambient luminosity, shown in \cref{fig:illumination_test}.
The detection rate is the ratio of successful wand detections to the total number of observations.
The marker can be robustly detected for illuminations up to $1000$ lux, which is the standard illumination of an office.

The blinking frequency of the blinking LED is controlled with a pulse-width modulation (PWM) signal from a microcontroller. 
The duty cycle of the PWM signal was set to 50\% to evenly space the on/off events for the event camera.

\begin{figure}[ht]
    \centering
    \includegraphics[width=0.7\linewidth]{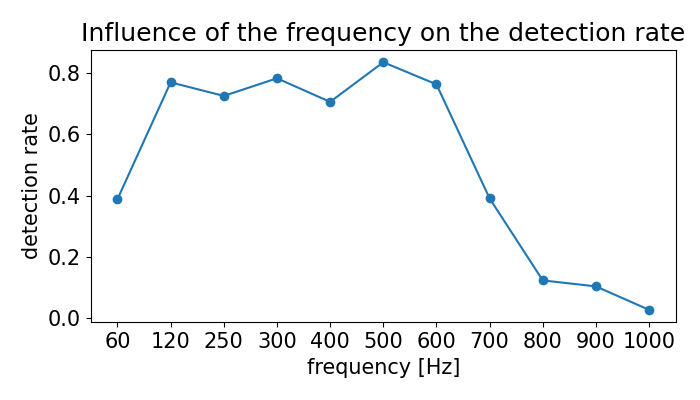}
    \caption{Influence of the frequency on the detection rate of the marker.}
    \label{fig:detec_rate_fq}
\end{figure}

As for the blinking frequency of the LEDs, we tuned it to obtain the best detection rate with the event cameras' default settings.
Intuitively, we want the highest possible frequencies in order to distinguish events generated from the blinking (high frequency) and events generated from the marker's movement (low frequency).
However, because of the refractory period of event-based cameras, the blinking frequency cannot reach the kHz levels achievable by the microcontroller.
In \cref{fig:detec_rate_fq}, we show that frequencies higher than 600 Hz lead to lower detection rates of the markers.

\subsection{Marker position extraction}\label{subsec:marker_position_extracter}

In this section, we will describe the extraction of the marker position for frame-based cameras in \cref{subsubsec:marker_position_extracter_frame} and for event-based cameras in \cref{subsubsec:marker_position_extracter_event}.

\subsubsection{Frame-based camera}\label{subsubsec:marker_position_extracter_frame}

\begin{figure}[h!]
  \centering
    \begin{subfigure}[b]{0.49\linewidth}
        \centering
        \includegraphics[width=\linewidth]{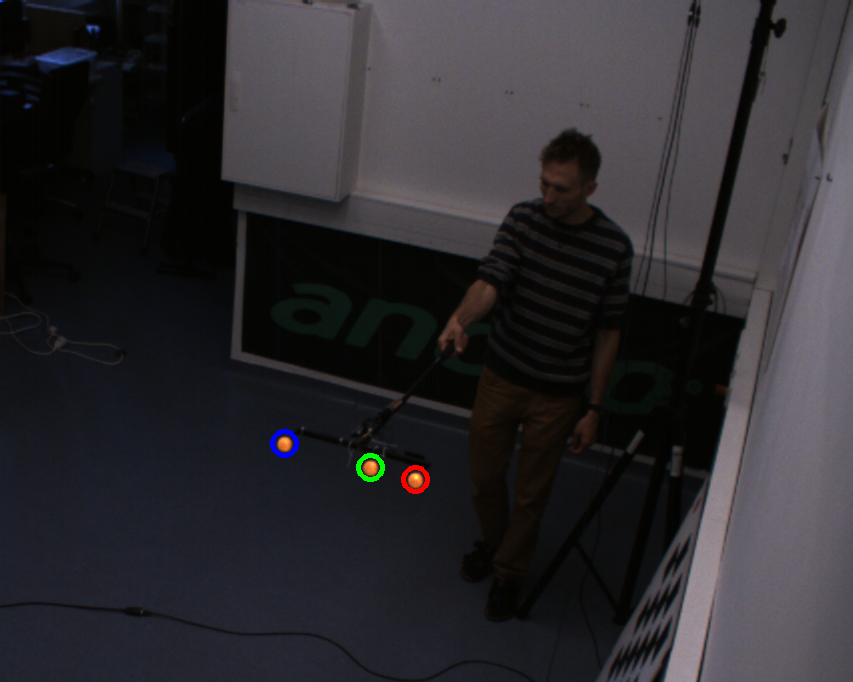}
        \caption{frame\_0}
    \end{subfigure}
    \hfill
    \begin{subfigure}[b]{0.49\linewidth}  
        \centering 
        \includegraphics[width=\linewidth]{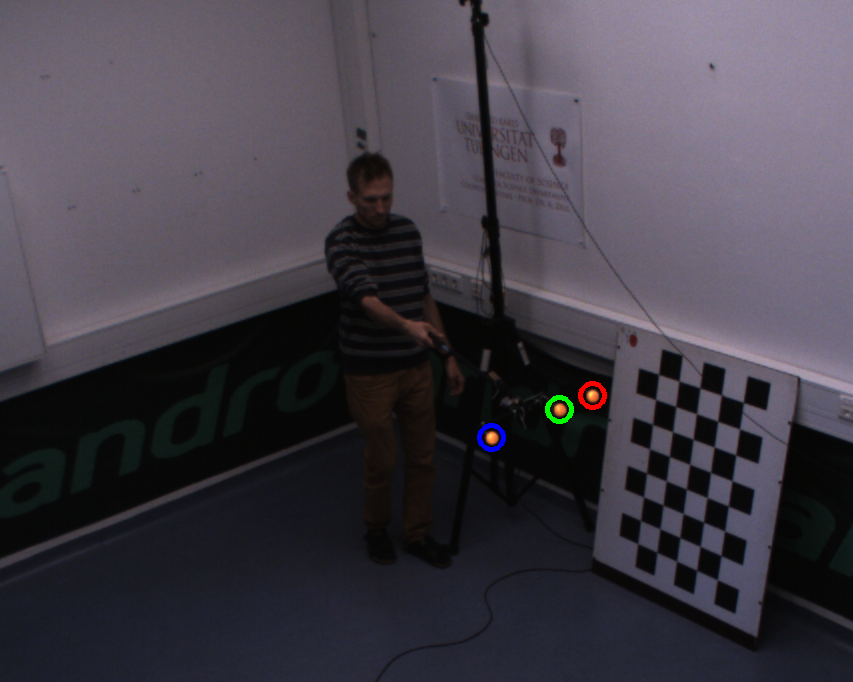}
        \caption{frame\_1}
    \end{subfigure}
    \hfill
    \begin{subfigure}[b]{0.49\linewidth}   
        \centering 
        \includegraphics[width=\linewidth]{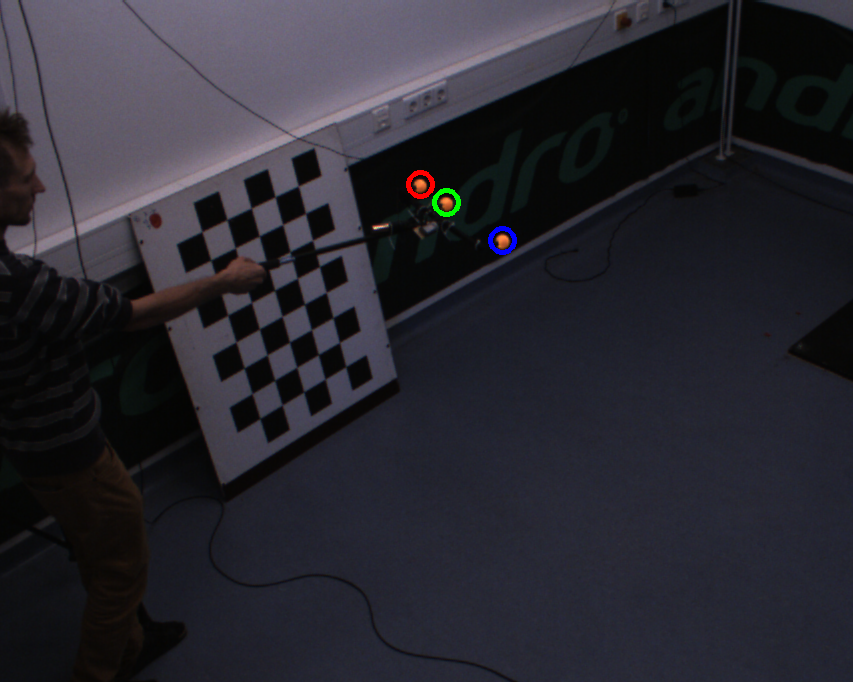}
        \caption{frame\_2}
    \end{subfigure}
    \hfill
    \begin{subfigure}[b]{0.49\linewidth}   
        \centering 
        \includegraphics[width=\linewidth]{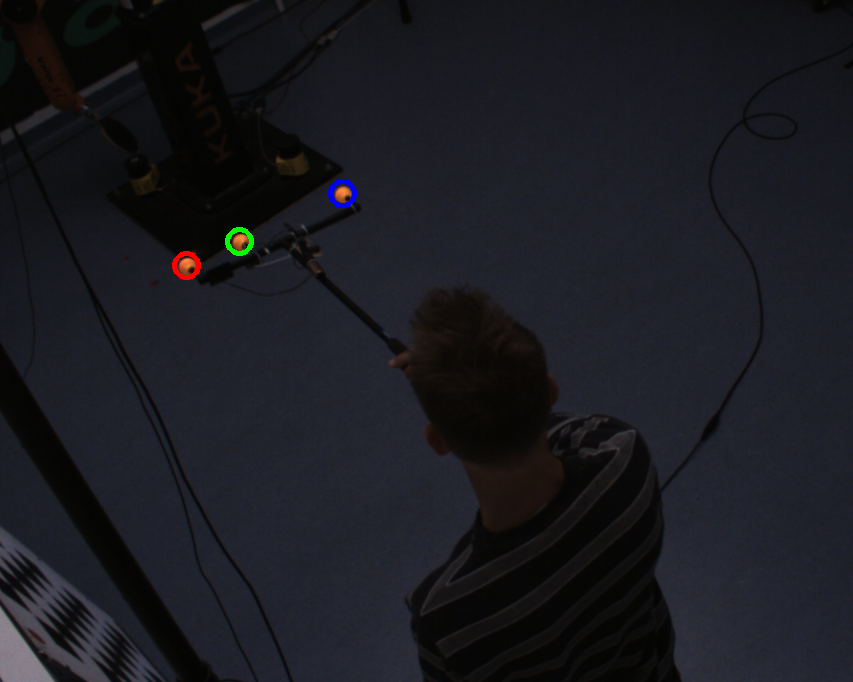}
        \caption{frame\_3}
    \end{subfigure}
    
  \caption{
    Wand observed with the frame-based cameras. The position of the balls is marked with the {\color{red}red}, {\color{green}green} and {\color{blue}blue} circles.
  }
  \label{fig:marker_position_frame}
\end{figure}

For frame-based cameras, we identify the markers using a combination of color segmentation and blob detection, using the \textit{OpenCV} implementation.
However, false positives can still occur.
To further filter out the detected blobs, we only keep blobs close to each other and aligned.
An example of a simultaneously detected wand in each of the four frame-based cameras is visualized in \cref{fig:marker_position_frame}.
It is important that the camera's exposure time is longer than a blinking period to avoid brightness inconsistencies in the marker's color.
We also keep the exposure as low as possible to make the blinking marker stand out compared to the darker background.

\subsubsection{Event-based camera}\label{subsubsec:marker_position_extracter_event}

By tuning the biases of the event cameras (high and low pass filter mostly), it is possible to filter out most of the events except those generated by the marker.

However, this approach is not appropriate for tasks that also involve capturing events generated by other movements.
Instead, we chose to extract the frequency of the generated events to filter out any frequencies that are not that of the blinking LEDs. 

\begin{figure}
  \centering
  \begin{subfigure}[b]{0.10\textwidth}
     \includegraphics[width=\textwidth]{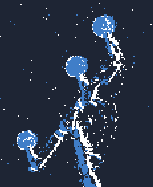}
     \caption{}
     \label{subfig:accumulated_events}
   \end{subfigure}
   \begin{subfigure}[b]{0.16\textwidth}
     \includegraphics[width=\textwidth]{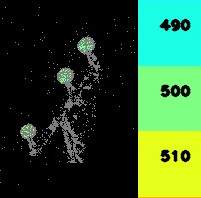}
     \caption{}
     \label{subfig:frequencycam_output}
   \end{subfigure}
   \begin{subfigure}[b]{0.10\textwidth}
     \includegraphics[width=\textwidth]{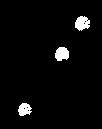}
     \caption{}
     \label{subfig:blob_output}
   \end{subfigure}
  \caption{(a) Accumulated events from the wand. (b) The detected frequencies of the marker positions by FrequencyCam~\cite{Pfrommer2022arxiv}. (c) Output of the blob detector.}
  \label{fig:event_marker_extraction}
\end{figure}

We rely on FrequencyCam~\cite{Pfrommer2022arxiv} for that purpose.
FrequencyCam takes an event stream as input and determines the frequency of every pixel in the field of view of the event-based camera.
A visualization of the output of FrequencyCam can be seen in \cref{subfig:frequencycam_output}.
We modified FrequencyCam to record only pixels corresponding to the frequency of our blinking LEDs, employed blob detection (as shown in \cref{subfig:blob_output}) for precise marker center determination, and exported both position data and timestamps.

\section{eWand}\label{sec:ewand}

\begin{figure}[h!]
  \centering
  \includegraphics[width=0.8\linewidth]{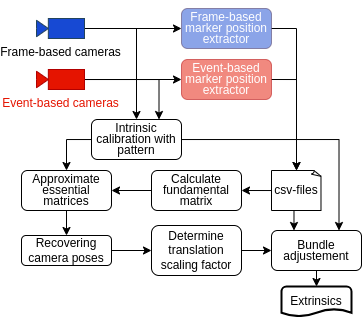}
  \caption{
    An overview of our proposed calibration method.
  }
  \label{fig:method}
\end{figure}

In this section, we provide a general overview of our calibration method, visualized in \cref{fig:method}.
We describe our calibration target in \cref{subsec:calibration_target}.
How we performed the intrinsic camera calibration is outlined in \cref{subsec:intrinsic_calib}.
In \cref{subsec:extrinsic_calib} we detail the extrinsic calibration.

\subsection{Calibration target}\label{subsec:calibration_target}

The wand is composed of three of our markers, described in \cref{sec:event_camera_marker}, mounted in a line onto a stick, as shown in \cref{fig:ewand}.
These spheres constitute the pattern's markers for the calibration. 
Marker size and spacing were chosen considering construction constraints and camera resolution.
The spheres are aligned such that it is possible to distinguish them by the distance ratio between them.
Indeed, the ratio is conserved regardless of the viewing angle.
This is necessary to have marker correspondence between all cameras.

The wand was 3D printed in one piece to ensure maximal accuracy of the markers' position.
However, because it was hand-built, it is still prone to inaccuracies ($\pm 2mm$).

\subsection{Intrinsic calibration}\label{subsec:intrinsic_calib}

In order to estimate the camera matrix $\mathbf{K}$ and the distortion coefficients $\bm{d}$, we need a 3D or 2D pattern with known geometry.
Multiple methods~\cite{Hartley2001}\cite{Bougnoux1998iccv}\cite{Hartley2003} are available, but for all of them, the 3D pattern requires more than three points.
The intrinsic parameters' initialization method~\cite{Zhang2000pami} requires planar patterns and cannot be applied to the wand, which is just a line.

Moreover, the inaccuracies in the marker positions of the wand, since it is self-manufactured, make it ill-suited for intrinsic calibration, which is very sensitive to bias and noise.
For these reasons, we do not rely on the wand for the intrinsic calibration.
We use the intrinsic parameters calculated with the \textit{calibrateCamera} function from OpenCV with the asymmetric circle pattern and reconstruction with E2VID\cite{Rebecq2019pami}.
Camera intrinsics remain constant and only require one-time calibration, unlike extrinsics, which must be recalculated whenever a camera is moved even slightly.
This is why the ease of use of eWand makes it preferable over other methods. 

\subsection{Extrinsic calibration}\label{subsec:extrinsic_calib}

For the extrinsic calibration, we start by recording images and events while moving the wand around.
After capturing the data from the event- and frame-based cameras, we extract the position of the markers, explained in \cref{subsec:marker_position_extracter}.

We then rely on bundle adjustment for extrinsic calibration.
It can be seen as an optimization problem that requires finding the best extrinsics and 3D marker positions to minimize the reprojection errors
\begin{equation}\label{eq:bundle_adjustment}
  argmin_{\mathbf{P}_j, \hat{\mathbf{X}}_{i}, \mathbf{K}, \mathbf{d}} \sum^n_{i=1} \sum^m_{j=1} (x_{i,j} - \hat{x}_{i,j}(\mathbf{P}_j, \hat{\mathbf{X}}_{i}, \mathbf{K}, \mathbf{d}))^2
\end{equation}
where $x_{i,j}$ is the $i\text{th}$ observed marker with camera $j$, 
$\hat{\mathbf{X}}_{i}$ is the estimated 3D position of the $i\text{th}$ marker observation, $\mathbf{P}_j$ is the projection matrix of the $j^\text{th}$ camera, and $\hat{x}_{i.j}$ is the prediction of $x_{i,j}$ using $\mathbf{P}_j$, $\hat{\mathbf{X}}_{i}$ and the intrinsic camera parameters $\mathbf{K}$ and $\bm{d}$.
The projection matrix is the product of the camera matrix, that we know, and of the camera pose, that we want to estimate.
We rely on Ceres~\cite{Agarwal_Ceres_Solver_2022} to perform the optimization part of the bundle adjustment.

Additional costs are added to the optimization problem to use the extra information we have from the wands' structure as a constraint. We assume here that index 0, 1 and 2 belong to the same wand.
\begin{equation} 
    \text{Linearity: } (\hat{\mathbf{X}}_{0} - \hat{\mathbf{X}}_{1}) \times (\hat{\mathbf{X}}_{1} - \hat{\mathbf{X}}_{2}) = 0
\end{equation}
\begin{equation}
    \text{Distance: } | \| \hat{\mathbf{X}}_{0} - \hat{\mathbf{X}}_{1}\| - l_{ref,0} | + |  \|\hat{\mathbf{X}}_{1} -\hat{\mathbf{X}}_{2}\| - l_{ref, 1} | = 0 \\
\end{equation}
where $l_{ref, 0}$ and $l_{ref, 1}$ are the real distances between the wand's markers.
Non-linear optimization problems, such as bundle adjustment, require a good initialization of the extrinsic parameters so as not to get stuck in the wrong local minimum.
To do so, we first need the fundamental matrix $\bm{F}_j$ of a pair of cameras.
The fundamental matrix is obtained using a RANSAC version of the 7-point algorithm.
Once we have $\bm{F}_j$, we can calculate the essential matrix $\bm{E}_j$ with the camera matrix of the two cameras $\bm{K}_j$ and $\bm{K}_k$ using the formula: $\bm{E}_j = (\bm{K}_k)^T \bm{F_j} \bm{K_j}$.
The essential matrix can then be decomposed into possible camera rotation $\bm{R}_j$ and translation $\bm{t}_j$, using SVD.
However, the estimated translation vector $\bm{t}_j$ is only defined up to scale, but since we know the distance between the markers on the wand, we can retrieve the correct scale $k$ using a golden-section search. 

\section{Experiment}\label{sec:experiment}

\subsection{Cameras}\label{subsec:cameras}

\begin{figure}[!htbp]
  \centering
  \includegraphics[width=0.7\linewidth]{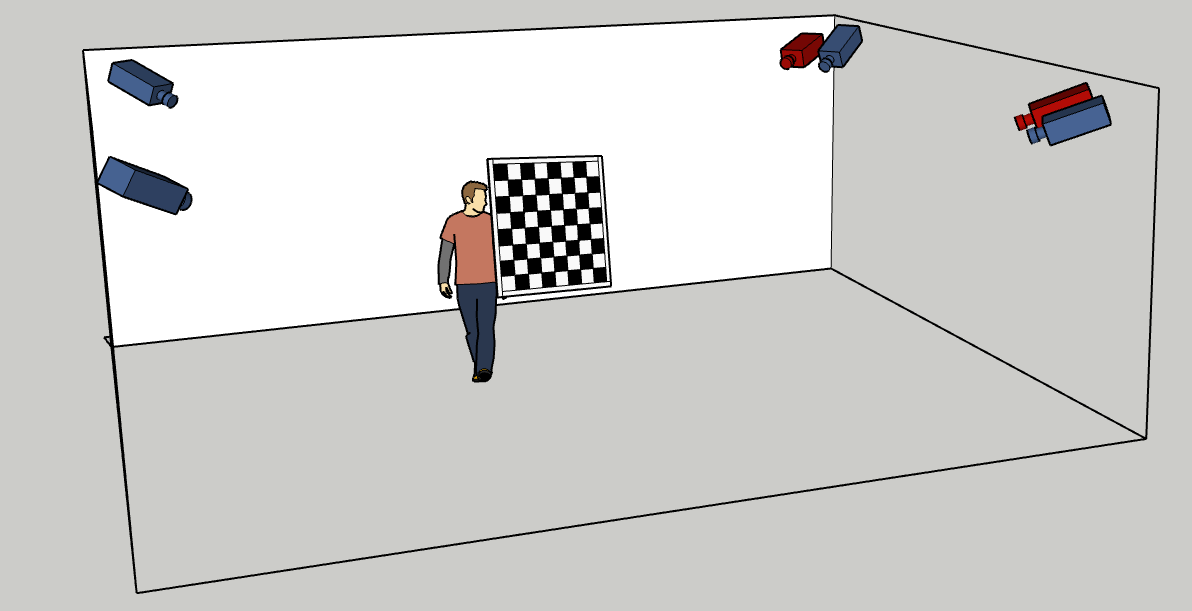}
  \caption{
    Our camera setup consisting of {\color{blue}four frame-based cameras (in blue)} and {\color{red}two event-based cameras (in red)} with baselines of $3$m to $5$m.
    Schematic is up to scale.
  }
  \label{fig:camera_setup}
\end{figure}

Our setup contains a total of four frame-based and two event-based cameras with baselines of $3$m to $5$m, organized as shown in \cref{fig:camera_setup} (schematic is up to scale).

Four FLIR Chameleon3 frame-based cameras (1280$\times$1024 pixels) are mounted in the room's corners.
As for the event-based cameras, two Prophesee EVK4 cameras (1280$\times$720 pixels) are placed at two corners of the room.
The event-based cameras are “paired” with a frame-based camera to have both type of data with comparable points of view.

The parameters of both types of cameras were set to record the best quality of data.
For the frame-based cameras, we want distinguishable ball outline compared to the rest of the scene.
For the event-based cameras, we want to avoid noise and capture all the marker blinks.
For that, we adjusted the trigger thresholds (defining the level of the illumination change necessary to trigger an event) to filter out noise but keep the events from the blinking markers.
\subsection{Time synchronization}\label{subsec:timesync}

One of the most critical aspects of multi-camera calibration is the time synchronization of the data capture.
Because the wand is constantly moving, a frame offset could lead to erroneous triangulation of the balls.
In our setup, we use a microcontroller to trigger the frame-based cameras and the primary event-based camera with $50$ Hz.
The clocks of the primary and secondary event-based camera are synchronized via a synchronization cable between them.
\begin{table*}[!ht]
    \centering
    \resizebox{\textwidth}{!}{
	\begin{tabular}{l|l|l|l|l|l|l}
		\textbf{Camera} & \textbf{BA eWand (our)} & \textbf{kalibr (circleboard)} & \textbf{kalibr (checkerboard)} & \textbf{BA circleboard} & \textbf{BA checkerboard} \\
		\toprule
        frame\_0 & $0.325 \pm 0.257$ & $0.274 \pm 0.060$ & $0.390 \pm 0.325$ & $0.218 \pm 0.219$ & $0.225 \pm 0.204$ \\ 
        frame\_1 & $0.320 \pm 0.271$ & $0.311 \pm 0.070$ & $0.398 \pm 0.335$ & $0.201 \pm 0.178$ & $0.201 \pm 0.185$ \\
        frame\_2 & $0.401 \pm 0.361$ & $0.168 \pm 0.072$ & $0.320 \pm 0.320$ & $0.178 \pm 0.178$ & $0.213 \pm 0.193$ \\
        frame\_3 & $0.396 \pm 0.389$ & $0.159 \pm 0.062$ & $0.301 \pm 0.243$ & $0.199 \pm 0.188$ & $0.180 \pm 0.165$ \\
        event\_0 & $0.603 \pm 0.448$ & $0.269 \pm 0.068$ & $0.513 \pm 0.396$ & $0.398 \pm 0.365$ & $0.459 \pm 0.363$ \\ 
        event\_1 & $0.550 \pm 0.431$ & $0.295 \pm 0.070$ & $0.515 \pm 0.405$ & $0.469 \pm 0.447$ & $0.402 \pm 0.370$ \\
		\hline
	\end{tabular}
    }
    \caption{Reprojection error (MAE) in pixels (mean and std.) for each calibration approach and camera, after the calibration.
             "BA" represents the bundle adjustment approach using OpenCV and Ceres, explained in \cref{subsec:calib_opencv_ceres}.}
    \label{tab:reprojection_errors}
\end{table*}
\subsection{Comparison}\label{subsec:comparison}
We compare the extrinsics estimated by our proposed calibration method, described in \cref{subsec:calib_wand} with two other calibration approaches.
On the one hand, we use the popular calibration framework kalibr~\cite{Rehder2016icra}, described in \cref{subsec:calib_kalibr}.
On the other hand, an intrinsic calibration method based on OpenCV~\cite{opencv_library} and an extrinsic calibration method using OpenCV and the optimization framework Ceres~\cite{Agarwal_Ceres_Solver_2022} to solve the calibration as a bundle adjustment problem, described in \cref{subsec:calib_opencv_ceres}.
In \cref{subsec:evaluation}, we describe our evaluation of this experiment.

\subsection{Calibration with kalibr}\label{subsec:calib_kalibr}

We performed the intrinsic and extrinsic calibration with one dataset for each calibration target, following~\cite{Muglikar2021cvprw}.

\subsection{Calibration via bundle adjustment with OpenCV/Ceres}\label{subsec:calib_opencv_ceres}

For the calibration with OpenCV and bundle adjustment, we chose to perform the intrinsic and extrinsic calibration separately and on two different data sets.

Doing the calibration in two stages (first intrinsics and afterward extrinsics) has the advantages that recordings can be used which are specifically targeted for one of the two tasks.
For the intrinsic calibration, the calibration target should be recorded in the cameras' whole field of view.
On the other hand, for the extrinsic calibration, the calibration target has to be observed by all cameras simultaneously, not able to cover the individual cameras' whole field of view.
We use two standard patterns: checkerboard and asymmetric circle grid.
We chose not to include AprilTags as they tend to have a poor detection rate after reconstruction from events~\cite{Muglikar2021cvprw}.
For the event-based cameras, E2VID~\cite{Rebecq2019pami} was used to generate reconstructed frames from the events.

\paragraph{Intrinsic calibration} For the intrinsic calibration, we used OpenCV functions to detect the calibration targets, followed by the \textit{cameraCalibrate} function.

\paragraph{Extrinsic calibration} To perform the extrinsic calibration, we first calculated an initial estimate of the extrinsics and run a bundle adjustment afterward.
For the initial estimate, we used \textit{solvePnP}, given the observed target for every camera to get the position and rotation of the target in the camera frame.
With these positions and rotation, we calculated the position and rotation of every camera with respect to the main camera.
These initial estimates were then used as the starting point for a bundle adjustment problem, where we used Ceres~\cite{Agarwal_Ceres_Solver_2022} to optimize the intrinsics and extrinsics of all the cameras.

\subsection{Calibration with the wand}\label{subsec:calib_wand}

We recorded a data set where we moved the wand around, covering the whole working space.
The blinking frequency of the LEDs were set to $500$ Hz.
Since the wand is easier to detect, we were able to cover a bigger space compared to the other calibration targets.
For the calibration we proceeded as explained in \cref{sec:ewand}.

\subsection{Evaluation}\label{subsec:evaluation}

For the evaluation we used two metrics: the commonly used reprojection error and a 3D position error.

\subsubsection{Reprojection errors}

\Cref{tab:reprojection_errors} lists the reprojection errors obtained after calibrating with each method.

It is important to note that the measurements used differ for eWand and the other methods.

As can be seen, our method achieves similar reprojection errors as the other methods.
Considering the standard deviation, the error ranges do overlap.
The lower reprojection error of kalibr with the circleboard might come from the fact that kalibr does some outlier rejection, while our proposed method and OpenCV do not.

\subsubsection{3D error}

\begin{figure}[!h]
    \centering
    \includegraphics[width=0.7\linewidth]{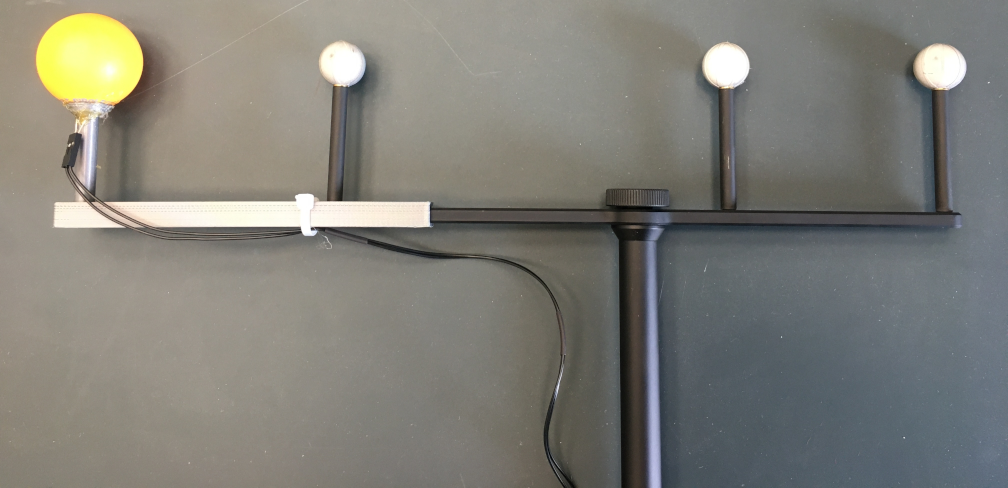}
    \caption{
    Modified OptiTrack wand used for benchmarking.
    }
    \label{fig:benchmarking}
\end{figure}
For the evaluation, we attached one of our custom marker to an OptiTrack wand, as shown in \cref{fig:benchmarking}.
The marker is the same as the one described in \cref{sec:event_camera_marker}.
We moved the OptiTrack wand with the attached marker within the workspace.
Taking the 2D observations of the marker in all the cameras, we calculate the 3D position by solving a bundle adjustment problem, similar to \cref{eq:bundle_adjustment}, where only the 3D position of the marker is optimized.
Finally, we compared the 3D position from the OptiTrack system, which we take as ground truth, to the ones we got from the bundle adjustment.
The transformation between the Optitrack frame and our setup's frame is calculated by matching both generated point clouds.
For the time synchronization of both, we determine the time offset by minimizing the error with the previously found transformation.
\begin{table}[!h]
    \centering
    \begin{tabular}{l|l} 
        \textbf{Method} & \textbf{3D positional error [mm] (mean and std)} \\
        \toprule 
        BA (checkerboard) & $47.0 \pm 35.8$\\
        \hline
        BA (circleboard) & $84.9 \pm 35.8$\\
        \hline
        kalibr (checkerboard) & $8.4 \pm 10.5$\\
        \hline
        kalibr (circleboard) & $8.6 \pm 10.4$\\
        \hline
        BA eWand (our) & $9.3 \pm 10.4$ \\ 
        \hline
    \end{tabular}
    \caption{
    The 3D position MAE [mm] (mean and std) of the different calibration methods.
    }
    \label{tab:3d_position_errors}
\end{table}
As we can see in \cref{tab:3d_position_errors}, the positional error of our proposed method (eWand) is very close to the results of kalibr.
Considering the standard deviation, the error ranges do overlap.
The bundle adjustment approach using OpenCV and Ceres performs a magnitude worse than kalibr and our approach.
While kalibr does some outlier rejection, the bundle adjustment approach does not.
The positional error with the checkerboard is the lowest in our experiments.
One possible explanation is that on the frame reconstructions from the event data, the checkerboard detection might perform better compared to the circleboard.

\section{Conclusion}\label{sec:conclusion}

Calibrating camera systems is crucial, but the standard method of using printed patterns cannot be applied to event-based cameras.
State-of-the-art methods include reconstructing frames or using active markers.
Furthermore, calibration with the widely used calibration targets becomes challenging for multi-camera setups with a wide baseline.
Our \textbf{eWand} method extends wand-based calibration to event-based cameras, using blinking LEDs to eliminate the need for frame reconstruction.
\textbf{eWand} offers a simpler way to calibrate extrinsic parameters of multiple cameras, maintaining accuracy with less effort.
While achieving good results in this first step, our bundle adjustment can be improved, e.g., by introducing outlier rejection.
In the future, we also plan to make an improved wand with better manufacturing, for estimating intrinsic parameters as well.
Overall, \textbf{eWand} aims to make setting up camera recording systems with frame- and event-based cameras more accessible and user-friendly.


\addtolength{\textheight}{-8cm}
\bibliographystyle{IEEEtran}
\bibliography{IEEEabrv,bibliography}

\end{document}